\documentclass[conference]{IEEEtran}
\IEEEoverridecommandlockouts

\usepackage{amsmath,amssymb}
\usepackage{graphicx}
\usepackage{booktabs}
\usepackage{multirow}
\usepackage{array}
\usepackage{cite}
\usepackage{url}
\usepackage{balance}
\usepackage{stfloats}
\usepackage{placeins}

\newcommand{\papertablefont}{\footnotesize}

\graphicspath{{./}{figures/}{Figures/}{images/}{Images/}{plots/}{Plots/}}

\title{Frequency Is Not Sensitivity: Identifying Safety-Sensitive Experts in Sparse MoE LLMs}

\author{
\IEEEauthorblockN{Md Nurul Absar Siddiky, Liuwan Zhu, and Yingfei Dong}
\IEEEauthorblockA{
Department of Electrical and Computer Engineering\\
University of Hawaii at Manoa, Honolulu, HI 96822, USA\\
\{msiddiky,liuwan,yingfei\}@hawaii.edu}
\thanks{This work has been submitted to the IEEE for possible publication. Copyright may be transferred without notice, after which this version may no longer be accessible.}
}

\begin{document}

\maketitle

\begin{abstract}
Suppressing a small set of routed experts can weaken the safety behavior of a sparse Mixture-of-Experts (MoE) 
language model without retraining. Which experts to suppress is therefore a security question, 
and the usual answer is activation frequency --- but frequency measures use, not influence. 
We test an alternative: router-gradient sensitivity, the sensitivity of the sequence loss to the gate weights 
that select an expert. Across five MoE architectures we rank experts by each signal on 500 benign and 
500 malicious prompts and measure refusal on 100 held-out malicious prompts, under two budgets: 
equal expert counts, and equal nominal malicious routing traffic (1\%--5\%). 
Under each of the two budgets, router-gradient selection reduces refusals more than activation in 24 of 25 conditions,
and more than a ten-trial random mean in all 25. The largest effect is in OLMoE, 
where refusals fall from 34 to 9 of 100 prompts (73.53\% relative) with no degraded outputs, 
indicating substantive compliance rather than broken generation. 
After matching expert counts in every layer, gradient selection still
produces greater refusal reduction than activation in 23 of 25 conditions,
with two ties. 
An exploratory cross-model analysis links larger malicious-versus-benign
concentration gaps to greater peak gradient effects ($\rho=0.90$;
exact two-sided $p=0.083$, $n=5$).
Together, the results support gradient selection under the tested budgets.
\end{abstract}


\section{Introduction}

Sparse Mixture-of-Experts (MoE) language models increase capacity by routing each token through a small subset of feed-forward experts~\cite{shazeer2017outrageously,lepikhin2021gshard,fedus2022review}. Because the router makes this choice explicitly, an MoE model exposes a discrete, per-token decision that can be observed and altered. Recent work exploits this: biasing the router so a small set of experts is never selected weakens refusal at inference, with no retraining~\cite{jiang2026sparsesafety,xu2026routehijack}. For openly released models this makes expert selection a security question: an attacker must know which experts to disable, and a defender which to protect.

This requires a signal that ranks experts by importance to safety behavior. Most MoE safety work ranks by \emph{activation frequency} --- how often malicious prompts route to an expert~\cite{lai2025safex,fayyaz2026steermoe,zhang2026raset}. But frequency measures \emph{use}, not \emph{influence}. An expert may be selected constantly yet matter little to any particular behavior, while a rarely selected one may be pivotal. \emph{Router-gradient sensitivity} provides a signal of influence: how strongly the sequence loss responds to the gate weights selecting an expert. The two signals need not rank experts alike, and neither is known in advance to give the better targets.

We address two research questions. \textbf{RQ1:} Do benign and malicious prompts differ in model-wide and within-layer routing concentration? \textbf{RQ2:} Does router-gradient selection identify experts whose suppression reduces refusals more than activation-based or random selection? We evaluate five instruction-tuned models whose routed-expert pools span 256 to 6{,}144 experts~\cite{jiang2024mixtral,qwen15moe,deepseekmoe,olmoe,qwen3} (Table~\ref{tab:models}), ranking experts on 500 benign and 500 malicious prompts and suppressing them on a disjoint set of 100, under two budget protocols that control for expert count and for routing traffic.

Our contributions are twofold. First, a five-architecture analysis of routing concentration and overlap, including an exploratory link between malicious-versus-benign concentration gaps and peak suppression effects. Second, held-out comparisons of gradient, activation, and random selection under complementary budgets, with an exact per-layer-count control showing that the gradient advantage is not explained by layer placement alone.

\section{Related Work}

\textbf{MoE routing and specialization.}
Sparse MoE research has emphasized conditional computation, capacity, and load balancing~\cite{shazeer2017outrageously,lepikhin2021gshard,fedus2022review}, later examining token-level specialization~\cite{xue2024openmoe}, interpretable architectures~\cite{yang2025moex}, steerable routing~\cite{bandarkar2025multilingual}, and activation-magnitude ``super experts'' whose ablation collapses general capability~\cite{su2026superexperts}. These studies motivate treating each expert, identified by its layer and in-layer index, as the unit of analysis; but routing frequency alone does not reveal how sensitive predictions are to the router parameters selecting it.

\textbf{Safety evaluation and internal interventions.}
AdvBench and JailbreakBench provide harmful-prompt benchmarks for refusal and jailbreak behavior~\cite{zou2023advbench,chao2024jailbreakbench}. StrongREJECT shows low-quality generations should not count as harmful compliance~\cite{souly2024strongreject}, so we separate restricted, compliant, and degraded outputs; dense-model work also finds low-dimensional refusal mechanisms~\cite{arditi2024refusal}, suggesting safety can depend on localized structure.

\textbf{MoE safety and expert-level interventions.}
Recent MoE-specific studies manipulate behavior-associated experts using activation-based signals: SAFEx uses stability-selected activation probabilities and masking~\cite{lai2025safex}; SteerMoE steers safety via activation-rate differences~\cite{fayyaz2026steermoe}; RASET fine-tunes safety-critical experts via contrastive activation sensitivity~\cite{zhang2026raset}. Input-only prompts can steer generation toward weakly aligned experts~\cite{fei2026misrouter,xu2026routehijack}, again selecting targets from activation patterns. Internal approaches include gate-guided ablation to remove safety behavior without retraining~\cite{wu2026gatebreaker} and masking-based search to flip refusals into compliance~\cite{jiang2026sparsesafety}.
Closest to our work, Zhao et al.~\cite{zhao2025backdoormoe} use gradient sensitivity, ablated against activation-based alternatives, to localize experts for backdoor injection rather than safety-behavior suppression.

Existing MoE safety work thus relies mainly on activation strength or masking. To our knowledge, this is the first study to compare router-gradient sensitivity against activation frequency as a selection criterion for expert suppression, under matched budgets and a random baseline.

\section{Methods}

\begin{table}[!t]
\caption{Analyzed routed expert spaces (always-active shared experts excluded from rankings and interventions).}
\label{tab:models}
\centering
\papertablefont
\setlength{\tabcolsep}{5pt}
\renewcommand{\arraystretch}{0.94}
\begin{tabular}{lrrrr}
\toprule
Model & $L$ & $E$ & Top-$k$ & $N$\\
\midrule
Mixtral-8x7B     & 32 & 8   & 2 & 256\\
Qwen1.5-MoE      & 24 & 60  & 4 & 1,440\\
DeepSeek-MoE     & 27 & 64  & 6 & 1,728\\
OLMoE-1B-7B      & 16 & 64  & 8 & 1,024\\
Qwen3-30B-A3B    & 48 & 128 & 8 & 6,144\\
\bottomrule
\end{tabular}
\end{table}

\subsection{Models, Analysis Sets, and Evaluation Sets}

Table~\ref{tab:models} reports the number of sparse layers $L$, routed experts per layer $E$, and the total number of routed experts $N=LE$; always-active shared experts are excluded. Prompts use each model's native chat template (Qwen3 in non-thinking mode~\cite{qwen3}). Each model's analysis set comprises 500 benign AEGIS2.0 prompts~\cite{ghosh2025aegis2} and 500 malicious AdvBench prompts~\cite{zou2023advbench}, used throughout for routing analysis, malicious-minus-benign scoring, ranking, and traffic calibration.

Behavioral evaluation uses two additional AEGIS2.0 sets~\cite{ghosh2025aegis2}, comprising 100 malicious and 100 benign prompts, with no overlap with the 500-benign/500-malicious analysis set. The malicious set evaluates the baseline and suppression based on gradient, activation, and random expert selection, whereas the benign set audits gradient-based suppression only. Responses are manually labeled as \emph{restricted} (refusal or safety redirection), \emph{compliant} (a substantive answer to the prompt), or \emph{degraded} (empty, incoherent, repetitive, or unrelated).\footnote{These labels are used only for behavioral evaluation and do not influence expert identification.}

\subsection{Activation and Router-Gradient Scores}

For prompt $p$, let $A^{(p)}_{l,e}$ denote the number of tokens routed to expert $e$ in sparse layer $l$. We normalize within each layer and average over the prompt set $\mathcal{P}_g$ for group $g\in\{B,M\}$ (benign/malicious):
\begin{equation}
\widetilde{A}^{(p)}_{l,e}
=
\frac{A^{(p)}_{l,e}}
{\sum_{j=1}^{E}A^{(p)}_{l,j}},
\qquad
\overline{A}^{\,g}_{l,e}
=
\frac{1}{|\mathcal{P}_g|}
\sum_{p\in\mathcal{P}_g}
\widetilde{A}^{(p)}_{l,e}.
\label{eq:activation}
\end{equation}

For gradient analysis, no parameters are updated; gradients are collected only for router-gate weights. For the chat-formatted input sequence $x^{(p)}_1,\ldots,x^{(p)}_{T_p}$, we use the causal-language-model sequence loss
\begin{equation}
\mathcal{L}^{(p)}
=
-\frac{1}{T_p-1}
\sum_{t=2}^{T_p}
\log
P_\theta\!\left(
x^{(p)}_t\mid x^{(p)}_{<t}
\right).
\label{eq:sequence_loss}
\end{equation}
The loss uses only the formatted analysis prompt, with standard causal shifting; generated responses are not appended.

For a router row $W^{\mathrm{gate}}_{l,e,:}\in\mathbb{R}^{d}$, where $d$ is the router-input hidden dimension (the number of weights in each row), the prompt-level gradient score is the mean absolute sequence-loss gradient:
\begin{equation}
G^{(p)}_{l,e}
=
\frac{1}{d}
\sum_{j=1}^{d}
\left|
\frac{\partial \mathcal{L}^{(p)}}
{\partial W^{\mathrm{gate}}_{l,e,j}}
\right|.
\label{eq:gradient}
\end{equation}

The absolute value prevents cancellation. Unlike activation frequency, $G^{(p)}_{l,e}$ measures local sensitivity of formatted-prompt token prediction to a router row. It is not a refusal objective or direct ``safety gradient''; suppression separately tests the behavior of experts selected by this signal.

For model $m$, let $\overline{G}^{\,M,(m)}_{l,e}$ and $\overline{G}^{\,B,(m)}_{l,e}$ be the averages of~\eqref{eq:gradient} over the 500 malicious and 500 benign analysis prompts, respectively. We define gradient malicious dominance as
\begin{equation}
D^{G,(m)}_{l,e}
=
\overline{G}^{\,M,(m)}_{l,e}
-
\overline{G}^{\,B,(m)}_{l,e}.
\label{eq:dominance}
\end{equation}
For the activation baseline, we analogously define $D^{A,(m)}_{l,e}=\overline{A}^{\,M,(m)}_{l,e}-\overline{A}^{\,B,(m)}_{l,e}$ from~\eqref{eq:activation}. Gradient and activation targets are ranked in descending order of their respective malicious-minus-benign scores.

Computing $G^{(p)}_{l,e}$ requires a backward pass per analysis prompt to obtain router-gate gradients, whereas activation frequency requires only a forward pass; the gradient signal therefore carries a higher one-time profiling cost during expert identification, though this cost is not incurred at inference time once experts are selected.

\subsection{Suppression Budgets and Evaluation}

For a selected set $\mathcal{S}$, malicious routing-traffic
share is $\tau(\mathcal{S})=
\sum_{(l,e)\in\mathcal{S}}\overline{A}^{\,M}_{l,e}/L$.
The denominator is
$\sum_{l,e}\overline{A}^{\,M}_{l,e}=L$, since each routed
layer's shares sum to one. To form a traffic-target set,
we accumulate these shares in the relevant ranked or random
order and stop before adding the first expert that would
exceed the target; that expert is excluded. 

We first select gradient-ranked expert sets corresponding
approximately to 1\%, 2\%, 3\%, 4\%, and 5\% of malicious
routing traffic, measured on the 500 malicious analysis
prompts. These sets contain 3/5/7/9/11 experts for Mixtral,
7/20/31/43/53 for Qwen1.5, 9/13/21/28/35 for DeepSeek,
5/9/13/17/21 for OLMoE, and 10/22/33/45/57 for Qwen3.
Each expert is identified by its layer and in-layer index.
Using these gradient sets as the reference, we construct two
complementary comparisons.

\textbf{Same-count comparison.}
For each gradient-selected set, the activation-based and
random-selection methods suppress the same number of experts.
Activation-based selection follows the activation-dominance
ranking, whereas random selection uses a random ordering.
This comparison controls the number of suppressed experts,
but not their cumulative routing traffic or layer-wise
distribution.

\textbf{Traffic-target comparison.}
Because the same-count comparison does not control cumulative
routing traffic, we additionally construct activation-based
and random expert sets corresponding approximately to the
same nominal traffic targets of 1\%--5\%, using the same
traffic definition and analysis prompts. Expert counts may
therefore differ across methods and random trials, while the
original gradient-selected sets remain unchanged. Layer-wise
distributions may still differ across methods. Because experts
are discrete, the achieved traffic shares may fall below the
nominal targets.

\textbf{Depth-matched activation control.}
Using the same 500-benign/500-malicious analysis rankings,
for each model--$K$ setting we select the highest
activation-dominance experts within each sparse layer,
subject to selecting exactly as many experts as the gradient
set contains in that layer. This matches total expert count
and the complete layer-count distribution, not only mean
depth. Expert overlap is allowed; evaluation prompts and
suppression settings remain unchanged. Routing traffic is
not additionally matched in this control.

\textbf{Intervention and evaluation.}
Selected experts receive a negative router-logit
bias before Top-$k$ selection, making them effectively
unavailable; shared experts remain unchanged. All methods
use the same 100 malicious evaluation prompts, greedy
decoding, and a maximum of 200 new tokens. Let $B_m$ be the number of restricted responses produced
by model $m$ without expert suppression, and let
$C_m^{(q)}$ be the corresponding count after expert
suppression at setting $q\in\{1,\ldots,5\}$, evaluated
on the same 100 malicious prompts. Method and protocol indices are omitted for brevity. Relative refusal
reduction is
\begin{equation}
R_m^{(q)}=100\,\frac{B_m-C_m^{(q)}}{B_m}.
\label{eq:reduction}
\end{equation}
Positive values indicate fewer restricted responses
after suppression; negative values indicate more.

Random selection uses ten independently seeded trials per
model and protocol, sampling without replacement while
retaining enough experts per layer for native Top-$k$
routing. Random refusal reduction is computed by replacing
$C_m^{(q)}$ in~\eqref{eq:reduction} with the mean restricted
count across the ten trials.

\section{Results and Discussion}

\subsection{Routing Concentration}

For RQ1, we compare model-wide concentration, within-layer concentration, and benign–malicious expert overlap.

\textbf{Overall concentration and overlap.}
Let $k_{90}^{g}$ denote the smallest number of ranked
experts covering 90\% of group $g$'s activation
mass. We report
$\Delta k_{90}=100(k_{90}^{M}-k_{90}^{B})/N$ in percentage
points, where negative values indicate stronger concentration
for malicious than benign prompts. Table~\ref{tab:activation} shows this pattern
in all five models, most strongly in OLMoE and DeepSeek.
Nevertheless, benign and malicious Top-10 sets share 4--7
experts, and Jensen--Shannon divergence between their
model-wide distributions $P_g(l,e)=\overline{A}^{\,g}_{l,e}/L$,
computed using base-2 logarithms, remains small (0.0074--0.0408). Thus, greater malicious concentration
coexists with substantial overlap in expert usage.

\textbf{Within-layer concentration.}
Entropy is computed per prompt from
$\widetilde{A}^{(p)}_{l,e}$ in~\eqref{eq:activation} as
$H_l^{(p)}=-\sum_e\widetilde{A}^{(p)}_{l,e}
\ln\widetilde{A}^{(p)}_{l,e}$ and averaged over prompts within
each group. For effective-expert counts, we first compute
$E_{\mathrm{eff},l}^{(p)}=e^{H_l^{(p)}}$ and then average
$E_{\mathrm{eff},l}^{(p)}/E$ over prompts and layers;
smaller values indicate more concentrated usage relative
to the available expert pool. Malicious ratios are lower
in four models, with the largest differences in OLMoE
and DeepSeek. Qwen1.5 is essentially unchanged, showing
that greater model-wide concentration does not imply
stronger concentration within every layer.

\begin{table}[!t]
\caption{Routing concentration and overlap. B/M:
benign/malicious; effective-expert ratios are averaged over
prompts and layers. Regions are illustrative examples, not
the largest benign--malicious differences; layer indices
are zero-based.}
\label{tab:activation}
\centering
\papertablefont
\setlength{\tabcolsep}{1.2pt}
\renewcommand{\arraystretch}{1.08}
\begin{tabular*}{\columnwidth}
{@{\extracolsep{\fill}}lrrrc@{}}
\toprule
Model & $\Delta k_{90}$ &
\shortstack{$E_{\mathrm{eff}}/E$\\B/M} &
\shortstack{Top-10\\overlap} &
\shortstack{Example\\layers}\\
\midrule
Mixtral  & $-1.95$ & .8999/.8900 & 4 & L8--15\\
Qwen1.5  & $-2.22$ & .7454/.7455 & 7 & L23\\
DeepSeek & $-6.94$ & .6144/.5889 & 6 & L12,25\\
OLMoE    & $-7.32$ & .6362/.6079 & 6 & L3,6,14\\
Qwen3    & $-3.52$ & .3479/.3448 & 7 & L6,18--19\\
\bottomrule
\end{tabular*}
\end{table}

The layer profiles reveal different concentration patterns
(Table~\ref{tab:activation}, Example layers). Mixtral shows slightly
more concentrated malicious routing over L8--15, with 6.906 effective
experts versus 7.002 for benign prompts. Qwen1.5's malicious
usage is slightly broader at L23 (40.2 versus 39.8 effective
experts). OLMoE shows lower malicious effective-expert counts at L3
and L14, alongside a shared minimum at L6. Qwen3 shows smaller, mixed group differences.

\subsection{Held-Out Suppression}

\begin{figure*}[!t]
\centering
\includegraphics[width=\textwidth]{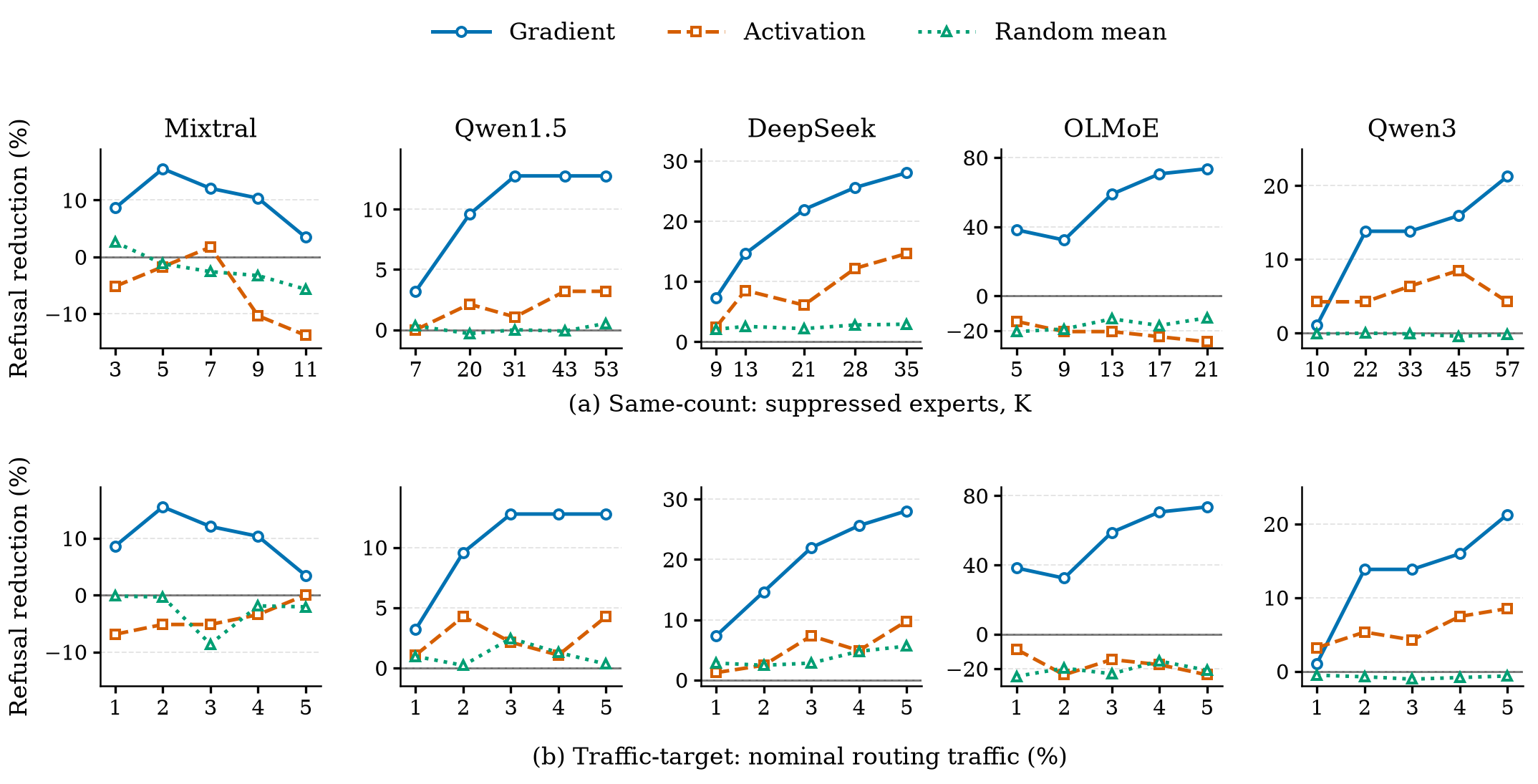}
\caption{Relative refusal reduction under (a) matched expert counts and
(b) nominal routing-traffic targets. Each column uses the same vertical
scale across rows; scales differ between models. Gradient sets and results
are reused; activation and random sets are selected separately for each
protocol. Random values use ten-trial mean restricted counts. Positive
values indicate fewer refusals and weaker safety; lines guide the eye.}
\label{fig:suppression_combined}
\end{figure*}

To answer RQ2, we test whether gradient-selected experts
produce greater refusal reduction than activation-selected
or randomly selected experts on held-out malicious prompts.
We first compare equal expert counts
(Fig.~\ref{fig:suppression_combined}(a)), then examine whether
the advantage persists under common nominal routing-traffic
targets (Fig.~\ref{fig:suppression_combined}(b)). Baseline restricted
counts $B_m$ are 58, 94, 82, 34, and 94 for Mixtral,
Qwen1.5, DeepSeek, OLMoE, and Qwen3, respectively, out of
100 prompts.

\textbf{Same-count comparison.}
Gradient selection yields greater refusal reduction than
activation in 24 of 25 model--$K$ conditions and exceeds
the random mean in all 25. The only exception is Qwen3
at $K=10$, where activation produces a 4.26\% reduction
versus 1.06\% for gradient selection. Since expert counts
are identical within each comparison, the gradient
advantage cannot be explained simply by suppressing
more experts.

The magnitude and progression of the effect differ across
models. OLMoE shows the largest relative reduction,
reaching 73.53\% at $K=21$: restricted responses fall
from 34 to 9. In contrast, activation and random
suppression increase OLMoE refusals at every setting.
DeepSeek's gradient reduction rises from 7.32\% to
28.05\%, and Qwen3 reaches 21.28\% at $K=57$.
Qwen1.5 plateaus at 12.77\% from $K=31$, while Mixtral
peaks at 15.52\% for $K=5$ and then declines.
Increasing the suppression budget therefore does not
necessarily increase refusal reduction.

\textbf{Traffic-target comparison.}
Equal expert counts do not ensure comparable routing
traffic. Figure~\ref{fig:suppression_combined}(b) therefore
compares new activation and random selections at the
same nominal 1\%--5\% traffic targets as the original
gradient sets. Gradient selection again exceeds
activation in 24 of 25 conditions and the random mean
in all 25. The exception remains Qwen3 at the smallest
setting: activation gives a 3.19\% reduction at 1\%,
compared with 1.06\% for gradient selection.
At the 5\% target, DeepSeek reductions are 28.05\%,
9.76\%, and 5.61\% for gradient, activation, and random
selection, respectively. For OLMoE, gradient suppression reduces refusals
while both controls increase them.

The gradient advantage therefore persists when experts are selected under common nominal routing-traffic targets rather than matched expert counts. Because only the activation and random sets are rebuilt to the traffic targets, this is a complementary check on a different confound, not an independent replication.

\textbf{Layer-position analysis.}
We normalize sparse-layer positions from 0\% (first) to 100\%
(last) and average over the 25 model--budget settings. Under
same-count budgets, gradient versus activation mean depths are
68.2\% versus 50.9\%, and deepest-third fractions are 56\%
versus 29\%; gradient sets are deeper in all 25 settings.
Activation stays shallow under traffic-target budgets
(48.7\%; 25.7\%). Settings are not independent replications
because budgets reuse models and overlapping sets.

\textbf{Depth-matched suppression.}
To test whether layer placement alone explains this advantage,
we compare gradient selection with the exact layer-matched
activation control (Table~\ref{tab:depth_matched}).
Gradient yields greater refusal reduction in 23 of 25
conditions, with ties for Qwen3 at $K=10$ (1.06\%) and
Mixtral at $K=11$ (3.45\%). At the largest budgets,
gradient versus depth-matched activation reductions are
28.05\% versus 4.88\% for DeepSeek and 73.53\% versus
$-29.41$\% for OLMoE. Thus, the observed gradient advantage
is not explained by layer placement alone, supporting its
value as a selection signal under the tested conditions.
This control does not isolate effects from differences
in selected routing traffic.

\begin{table}[!t]
\caption{Depth-matched activation refusal reduction $R$ (\%).
Settings $q=1,\ldots,5$ use each model's increasing $K$
sequence specified in Methods, not matched traffic targets.}
\label{tab:depth_matched}
\centering
\papertablefont
\setlength{\tabcolsep}{2pt}
\renewcommand{\arraystretch}{1.0}
\begin{tabular*}{\columnwidth}{@{\extracolsep{\fill}}lrrrrr@{}}
\toprule
Model & $q=1$ & $q=2$ & $q=3$ & $q=4$ & $q=5$\\
\midrule
Mixtral  & $-5.17$ & 6.90 & 6.90 & 1.72 & 3.45\\
Qwen1.5  & 0.00 & 1.06 & 1.06 & 1.06 & 2.13\\
DeepSeek & 2.44 & 7.32 & 6.10 & 7.32 & 4.88\\
OLMoE    & $-20.59$ & $-17.65$ & $-29.41$ & $-23.53$ & $-29.41$\\
Qwen3    & 1.06 & 0.00 & 1.06 & 0.00 & 1.06\\
\bottomrule
\end{tabular*}
\end{table}

\textbf{Connecting RQ1 to RQ2.}
Across the five models, the malicious-versus-benign concentration
gap $-\Delta k_{90}$ (Table~\ref{tab:activation}) is positively
associated with peak gradient-induced refusal reduction over
the five tested budgets (Fig.~\ref{fig:suppression_combined}):
Spearman $\rho=0.90$. An exact two-sided permutation test,
enumerating all $5!=120$ assignments of the five peak effects
to models, gives $p=10/120=0.083$ by counting correlations
with $|\rho|\geq0.90$. Thus, models with larger concentration
gaps tend to exhibit larger peak effects. This post hoc
association is suggestive, not evidence that concentration
causes susceptibility. It involves only five architectures,
uses maxima across tested budgets, and compares relative
reductions normalized by differing baseline refusal counts.

All three experiments produce no degraded outputs, and the benign
gradient-suppression audit yields no restricted or degraded responses.
These checks do not establish preserved accuracy, instruction following,
or general capability, but they do indicate that the refusal reductions
reflect substantive compliance rather than unusable generations,
reinforcing the depth-matched result.

\FloatBarrier

\section{Conclusion}

This study shows that activation frequency is not a consistently reliable proxy for suppression-induced behavioral effects in sparse MoE language models. Across five architectures, malicious inputs exhibit greater model-wide routing concentration despite substantial expert overlap with benign inputs, and router-gradient sensitivity consistently identifies suppression targets that reduce refusals more effectively than activation-based or random selection, with the largest effect a 73.53\% relative reduction in OLMoE. With exact per-layer counts matched, gradient selection exceeds activation in 23 of 25 conditions and ties in two, supporting an advantage beyond layer placement alone. The exploratory association between concentration gaps and peak gradient effects connects routing structure to behavioral susceptibility, but requires validation across more models. We intend these findings to support the identification and hardening of safety-sensitive experts in MoE models, not to enable misuse. The mechanism behind the non-monotonic effects observed at larger suppression budgets, along with general model capability preservation, remain topics for future study.

\balance

\end{document}